\documentclass[runningheads]{llncs}

\usepackage{accv}

\usepackage{accvabbrv}

\usepackage{graphicx}
\usepackage{booktabs}

\usepackage[accsupp]{axessibility}  

\usepackage[pagebackref,breaklinks,colorlinks,citecolor=accvblue]{hyperref}

\usepackage{orcidlink}

\begin{document}

\title{Underground Mapping and Localization Based on Ground-Penetrating Radar} 

\titlerunning{Abbreviated paper title}

\author{Jinchang Zhang\inst{1} \and
Guoyu Lu\inst{1} }

\authorrunning{F.~Author et al.}

\institute{Intelligent Vision and Sensing Lab, University of Georgia, USA }
\maketitle

\begin{abstract}
3D object reconstruction based on deep neural networks has gained increasing attention in recent years. However, 3D reconstruction of underground objects to generate point cloud maps remains a challenge. Ground Penetrating Radar (GPR) is one of the most powerful and extensively used tools for detecting and locating underground objects such as plant root systems and pipelines, with its cost-effectiveness and continuously evolving technology. This paper introduces a parabolic signal detection network based on deep convolutional neural networks, utilizing B-scan images from GPR sensors. The detected keypoints can aid in accurately fitting parabolic curves used to interpret the original GPR B-scan images as cross-sections of the object model. Additionally, a multi-task point cloud network was designed to perform both point cloud segmentation and completion simultaneously, filling in sparse point cloud maps. For unknown locations, GPR A-scan data can be used to match corresponding A-scan data in the constructed map, pinpointing the position to verify the accuracy of the map construction by the model. Experimental results demonstrate the effectiveness of our method.
  \keywords{3D Reconstruction \and Ground Penetrating Radar }
\end{abstract}

\section{Introduction}
\vspace{-1mm}
Ground-Penetrating Radar (GPR) is an ever-evolving, reliable, and effective technology for near-surface sensing. It is widely employed as a non-destructive imaging technique across various fields such as geological surveying \cite{wang2018velocity}, damage assessment \cite{benedetto2017overview} \cite{fernandes2017laboratory}, concrete scanning \cite{
pashoutani2020ground}, and detecting and localizing underground structures \cite{ 
lei2020underground}\cite{rhee2021study} including utility pipes, soil, and rebars. Owing to its exceptional capabilities, GPR has become a critical tool in remote sensing applications, aiding civil engineers and geophysicists. It uses a wave propagation technique to emit polarized high-frequency electromagnetic (EM) waves into the subsurface, which, upon encountering material changes, are reflected back and recorded. These reflections are processed and displayed as A-scan signals, while movement across a line perpendicular to underground structures creates B-scans, showing hyperbolic features indicative of object locations.

Modern localization systems, which often rely on visual and laser sensors, face challenges due to environmental variations like weather, lighting, and seasonal changes, affecting their performance. While laser scanners are less affected by light variations, they still struggle with issues such as weather and obstructions. GPR's primary advantage for localization tasks lies in its resilience to dynamic surface conditions, offering reliable mapping of subsurface data for precise localization. Unlike visual and LiDAR systems, which are sensitive to frequent environmental changes and require constant updates, GPR detects relatively stable subsurface features. This makes GPR a robust and efficient localization solution, overcoming the limitations of traditional visual and laser-based systems. Additionally, mapping and localization based on above-ground scenes rely on distinctive textures, such as corners. For texture-less scenes (e.g., snow and empty fields) or scenes with repetitive textures (e.g., forests and grasslands), mapping and localization accuracy is greatly affected.
To employ GPR for mapping and localization, we use a robot (HK1500) to tow a GPR cart (Leica DS2000) for data collection. SLAM systems are utilized to estimate the robot's pose during data collection. For B-scan data, we use ParNet to identify keypoints within the B-scans and fit parabolic equations to calculate the vertices, which represent the top contours of underground objects. Using the estimated poses, we stitch together a series of collected GPR images. However, the underground point cloud generated from GPR data is often sparse, necessitating point cloud completion. Due to the complexity of underground scenes, point cloud completion methods may cause closely spaced objects to merge, leading to inaccuracies. To address this, we designed GPRNet to simultaneously perform point cloud segmentation and completion, generating a dense underground 3D point cloud, which fulfills the mapping task. For localization, we leverage GPR A-scan data. For unknown locations, by collecting A-scan data and integrating map positions from previous data collections, we use NetVLAD to extract features from A-scans. These features are then matched with those from known positions in the map. For unknown locations, NetVLAD is used again to extract features and match them with existing data. Localization is achieved based on the best match, offering a precise and efficient solution for both mapping and localizing unknown points.

In summary, our main contributions include the following:
We introduce ParNet, which detects key points in B-scan data, fits parabolic equations, and determines the vertices that correspond to the depth of underground objects in cross-sections. By using poses calculated from a ground-based monocular SLAM system, we stitch multiple B-scans together to form a sparse point cloud.
For sparse point clouds in complex underground scenarios, we propose a multi-task point cloud network that simultaneously performs completion and segmentation. This network inputs sparse underground point clouds and outputs dense ones.
For established point cloud maps, we utilize A-scans for localization. By matching A-scan data from unknown locations with previously collected A-scan data, we can determine the coordinates of unknown locations within the point cloud.
\vspace{-1mm}
\section{RELATED WORK}
\vspace{-1mm}
\textbf{Mapping:}
\cite{Newcombe2011,Bloesch2018,whelan2015real} demonstrate the potential of achieving dense scene reconstruction by integrating per-pixel depth with RGB images. Depth information derived from raw images via deep learning can be seamlessly incorporated into SLAM systems, augmenting traditional frameworks with detailed depth maps. While supervised learning approaches \cite{eigen2014depth, ummenhofer2017demon, karsch2012depth} often outperform conventional methods, their reliance on labeled data for training limits their ability to generalize across new scenes. In contrast, recent advances have shifted towards unsupervised methods, reframing depth prediction as a problem of novel view synthesis. For instance, \cite{garg2016unsupervised} and \cite{godard2017unsupervised} employ image reconstruction loss to generate disparity maps, enforcing consistency between left and right images through a self-supervised approach, thus enabling end-to-end recovery of depth parameters. These dense depth maps offer solutions to some of the limitations inherent in classical monocular SLAM systems.
In terms of voxel representation, works such as \cite{ji2017surfacenet} and \cite{paschalidou2018raynet} explore multi-view reconstruction for volumetric scene and object representation. SurfaceNet \cite{ji2017surfacenet}, for example, predicts the confidence of a voxel to determine its surface presence, effectively reconstructing the scene's 2D surface. \cite{paschalidou2018raynet} reconstructs scene geometry by extracting view-invariant features and applying geometric constraints. There has also been significant progress in generating high-resolution 3D volumetric models, as seen in \cite{hane2017hierarchical} and \cite{tatarchenko2017octree}. However, in complex environments, geometric mapping alone is often insufficient; understanding the semantic information of objects within the scene is equally important. Semantic mapping bridges object classification and material composition with geometric structure, enriching maps with meaning beyond mere geometry. By integrating semantic information, mobile agents gain a deeper understanding of their environment, enabling higher autonomy and an expanded range of functionalities.



\begin{figure}[t]
\begin{center}
\includegraphics[width=8cm, height=4cm]{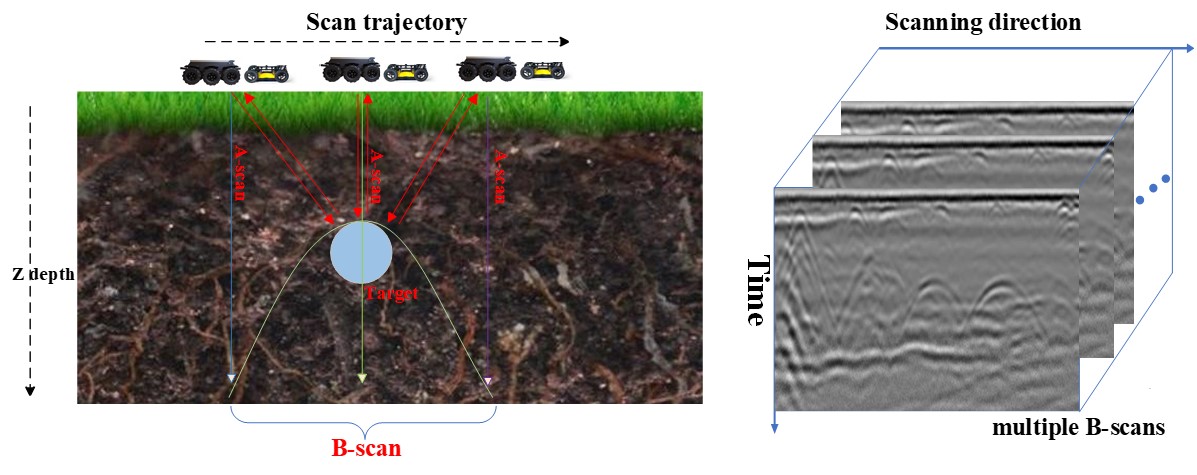}
\end{center}
\vspace{-7mm}
\caption{GPR sensing principles. Left: GPR emits EM radio wave pulses into the ground via an antenna. Pulses reflect back when hitting objects with different electromagnetic properties. Right:  Multiple B-scans along the moving direction can represent the sparse 3D underground object shape. }
\vspace{-7mm}
\label{realword}
\end{figure}

\begin{figure}[t]
\begin{center}
\includegraphics[width=7cm, height=4cm]{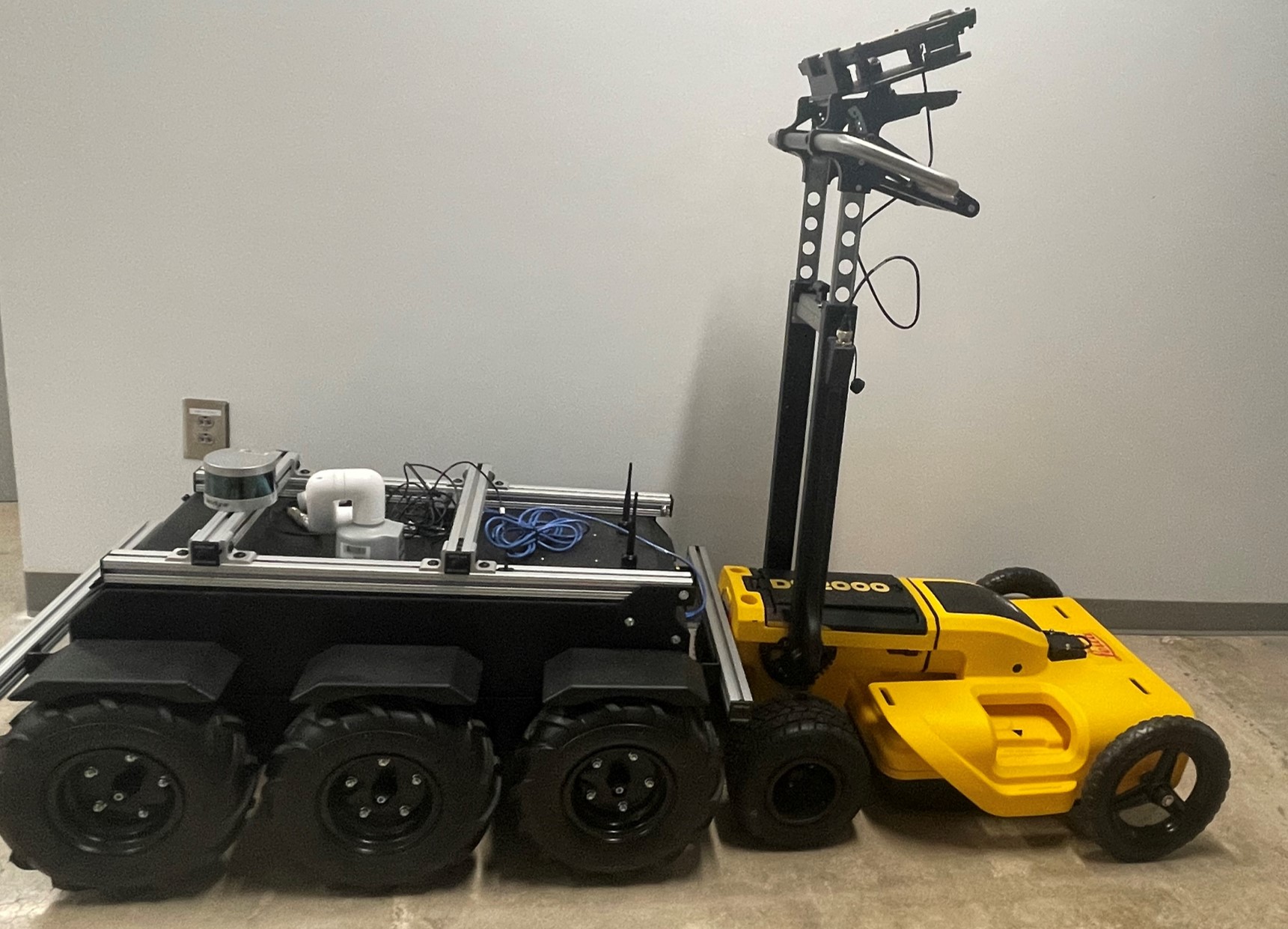}
\end{center}
\vspace{-5mm}
\caption{Omnidirectional robot for GPR data collection, where the robot (HK1500) tows a GPR cart (Leica DS2000).}
\vspace{-8mm}
\label{cart}
\end{figure}

\begin{figure*}[t]
\begin{center}
\includegraphics[width=12cm, height=4cm]{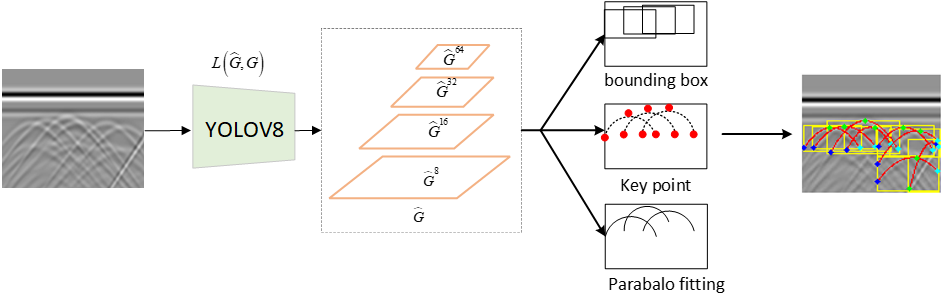}
\end{center}
\vspace{-7mm}
\caption{Parnet uses YOLOv8 architecture trained using the multi-task loss to map an GPR Bscan image to a set of output grids $\hat{\mathbf{G}}$ containing the predicted keypoints and bounding boxes. The equation of a hyperbola can be derived from its keypoints.}
\label{parnet}
\vspace{-6mm}
\end{figure*}
\textbf{GPR Interpretation: }
As the mathematical foundation behind traditional GPR imaging tools, the process known as migration is instrumental in converting unfocused raw B-scan radar graph data into focused targets. Migration techniques can broadly be categorized into Kirchhoff migration \cite{smitha2016kirchhoff}, the phase-shift migration \cite{gazdag1978wave}, the finite difference method \cite{claerbout1985downward}, and the back-projection algorithm \cite{demirci2012study}. While traditional methods facilitate the localization of buried objects, results reveal that they fail to provide detailed information regarding the shape and extent of these targets. Additionally, machine learning-based methods for interpreting GPR data, such as the Hough transform \cite{al2000automatic} and SVM \cite{ozkaya2020gpr}, have been extensively studied. Nevertheless, in the face of complex underground scenarios, machine learning techniques fall short in addressing complex curves and diverse signals. Contrary to these methods, an increasing array of deep neural network (DNN)-based applications are being employed for GPR data interpretation \cite{liu2020detection}. With the rapid development of object detection, \cite{feng2020gpr,lu20223d} utilizes detection networks to extract pertinent features from GPR images for underground object detection. Concurrently, \cite{feng2020gpr} pioneered the application of an enhanced Faster R-CNN-based framework to pinpoint buried objects, subsequently leveraging a depth estimation network to ascertain the depth values of the detected objects. Additionally, \cite{kaur2015automated} facilitates the localization of buried objects by identifying the positions of hyperbolas in GPR images. However, due to the MLP model's limitations in effectively capturing the characteristics of hyperbolas, the accuracy in pinpointing their apexes is compromised. This paper introduces the use of a keypoint detection network that excels in identifying the critical points of each hyperbola. By precisely fitting the hyperbolas through these keypoints, we are able to accurately locate their corresponding apexes. 

\vspace{-2mm}
\section{GPR-based Mapping and Localization}
\vspace{-1mm}
\subsection{Robotic GPR Data Collection}
\vspace{-1mm}
\label{data}

A GPR detects underground objects by emitting polarized pulses from its antenna, which reflect back upon encountering materials with different electromagnetic properties. The system captures the travel time of these echoes to create an A-scan, encapsulating signal strength and timing. Moving along a predefined path, the GPR compiles these A-scans into a B-scan, offering a two-dimensional underground map, illustrated on the left side of Fig. \ref{realword}. When operated across a survey grid, incorporating both horizontal and vertical trajectories, the GPR aggregates multiple B-scans to form a sparse three-dimensional representation of subsurface features, showcased on the right side of Fig. \ref{realword}.

As shown in Fig. \ref{cart}, we utilize the HK1500 robot to tow ground-penetrating radar for underground data collection. The HK1500 is a 6-wheel drive robot equipped with six heavy-duty motors and 13-inch tiller tires. This allows it to move forward, backward, and sideways, following a grid pattern without needing to rotate. Additionally, multiple 80/20 rails on the platform serve as attachment points for payloads. The rugged all-terrain design enables the HK1500 to carry cargo further and over rough terrain. A Leica DS2000 ground-penetrating radar is chosen, and the GPR antenna is mounted on a cart, which is towed by the HK1500 to collect data. A camera is mounted at the front of the robot, and a six-axis inertial measurement unit (IMU) embedded in the camera provides pose estimation together with the camera. The robot carries a rechargeable battery and a high-level controller to power the system and synchronize the pose data with the GPR scanning data.

\vspace{-1mm}
 \subsection{Parabola Keypoint Detection}
 \label{para}
 \vspace{-1mm}
The study by \cite{lu20223d} introduces a dual-network approach for parabola key point identification in images: a deep CNN predicts keypoints via target heatmaps, and another network fits parabola parameters. Despite its innovation, the method faces challenges: (1) Keypoint accuracy is limited by heatmap resolution, with larger heatmaps increasing processing demands. (2) Close keypoints within the same category may merge on the heatmap, leading to potential keypoint loss, a significant issue in b-scan data with adjacent parabolas. (3) The method is prone to overfitting and struggles with accurate parabola fitting in complex scenarios. To overcome these drawbacks, our network detects parabolas in B-scan data using bounding boxes and identifies three keypoints (the vertex and endpoints) within each box. By regressing these points, the parabola's equation is accurately determined. This approach sidesteps heatmap limitations, ensuring precise parabola shape and position identification, and improves robustness in images with overlapping parabolas by relying on local features for keypoint detection and equation calculation, mitigating the issues of direct regression error and overfitting.

\begin{figure*}[t]
\begin{center}
\includegraphics[width=12cm, height=8cm]{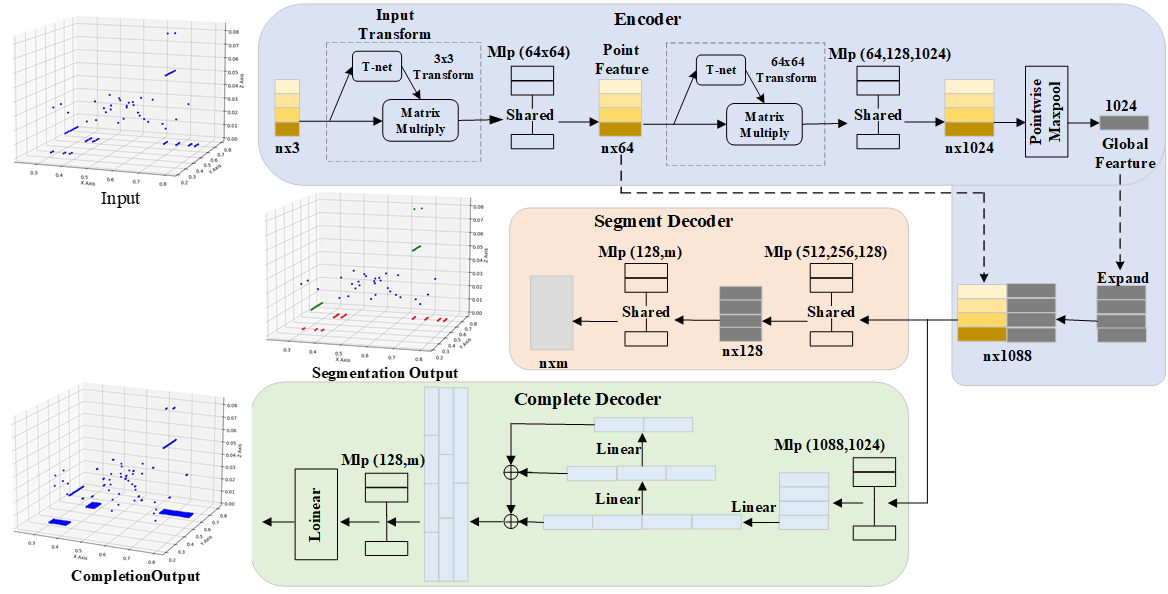}
\end{center}
\vspace{-7mm}
\caption{The GPRNet multi-task framework takes sparse point cloud data as input and simultaneously outputs segmentation results and the completed point cloud.}
\vspace{-5mm}
\label{Depth Methodology}
\vspace{-2mm}
\end{figure*}
Figure~\ref{parnet} illustrates the Parnet pipeline.The keypoint matching model references Kapao \cite{mcnally2022rethinking}. Employing a deep convolutional neural network $\mathcal{N}$ to process a GPR B-scan image $\mathbf{I}\in\mathbb{R}^{h\times w\times 3}$ into four output grids $\hat{\mathbf{G}} = \{\hat{\mathcal{G}}^s | s\in\{8, 16, 32, 64\}\}$, each representing object predictions $\hat{\mathbf{O}}$ with dimensions $\hat{\mathcal{G}}^s\in\mathbb{R}^{\frac{h}{s}\times \frac{w}{s}\times N_a \times N_o}$, $\text{anchor boxes } \mathbf{A}_s = (A_{w}, A_{h})$, the intermediate bounding box $\hat{t'} = (\hat{t'}_x, \hat{t'}_y, \hat{t'}_w, \hat{t'}_h)$. $\mathcal{N}$ is a YOLOv8 feature extractor. Building on~\cite{mcnally2022rethinking}, the intermediate bounding box $\hat{t} = (\hat{t}_x, \hat{t}_y, \hat{t}_w, \hat{t}_h)$ of an object is predicted using grid coordinates, relative to the origin of the grid cell $(i, j)$:
\vspace{-2mm}
\begin{equation}
\begin{array}{l}
\widehat{t}_x = 2\sigma \left( \widehat{t}_x' \right) - 0.5, \quad \widehat{t}_y = 2\sigma \left( \widehat{t}_y' \right) - 0.5 \\
\widehat{t}_w = \frac{A_w}{s} \left( 2\sigma \left( \widehat{t}_w' \right) \right)^2, \quad \widehat{t}_h = \frac{A_h}{s} \left( 2\sigma \left( \widehat{t}_h' \right) \right)^2
\end{array}
\end{equation}

Keypoints $\hat{v}$ are predicted in the grid coordinates and relative to the grid cell origin $(i, j)$ using:
\vspace{-3mm}
\begin{equation}
\begin{array}{l}
\widehat{v}_{xk} = \frac{A_w}{s} \left( 4\sigma \left( \widehat{v}_{xk}' \right) - 2 \right), \quad \widehat{v}_{yk} = \frac{A_w}{s} \left( 4\sigma \left( \widehat{v}_{yk}' \right) - 2 \right)
\vspace{-2mm}
\end{array}
\end{equation}

A target grid set $\mathbf{G}$ is established, utilizing a multi-task loss $\mathcal{L}(\hat{\mathbf{G}}, \mathbf{G})$ to train on object presence ($\mathcal{L}{obj}$), intermediate bounding boxes ($\mathcal{L}{box}$), keypoints ($\mathcal{L}{kps}$), and parabola fitting ($\mathcal{L}_{pra}$). The calculation of these loss components for an individual image is as follows:
\vspace{-1mm}
\begin{equation}
    \mathcal{L}_{obj} = \sum_s \frac{\omega_s}{n(G^s)}\sum_{G^s}\mathrm{BCE}(\hat{p}_o, p_o \cdot \mathrm{IoU}(\hat{\mathbf{t}}, \mathbf{t}))
\end{equation}
\vspace{-2mm}
\begin{equation}
    \mathcal{L}_{box} = \sum_s \frac{1}{n(\mathcal{O} \in G^s)}\sum_{\mathcal{O} \in G^s}1 - \mathrm{IoU}(\hat{\mathbf{t}}, \mathbf{t})
\end{equation}
\vspace{-2mm}
\begin{equation}
    \mathcal{L}_{kps} = \sum_s \frac{1}{n(\mathcal{O}^p \in G^s)}\sum_{\mathcal{O}^p \in G^s} \sum_{k=1}^K \delta(\nu_k > 0)\|\hat{\mathbf{v}}_k - \mathbf{v}_k\|_2
\end{equation}

\noindent where $\omega_s$ is the grid weighting, $\mathrm{BCE}$ is the binary cross-entropy, and $\nu_k$ represents the keypoint visibility flags.


 PraNet is capable of detecting three keypoints $(x_{left}, y_{left})$, $(x_{right}, y_{right})$, $(x_{top}, y_{top})$ on each parabola and calculating the corresponding parabolic equation parameters for comparison with the ground truth. $Par(x) = ax^2 + bx + c$
Therefore, the parabola fitting loss is optimized by L1 at each detected parabola center as:
\vspace{-1mm}
\begin{equation}
    \mathcal{L}_{pra} = \sum_s \frac{1}{n}\sum_{G^s} \left| {\widehat a - a} \right| + \left| {\widehat b - b} \right| + \left| {\widehat c - c} \right|
    \vspace{-1mm}
\end{equation}
The total loss, $\mathcal{L}$, is the weighted sum of the individual loss components.
\vspace{-2mm}
\begin{equation}
\mathcal{L} = \lambda_{obj}\mathcal{L}_{obj} + \lambda_{box}\mathcal{L}_{box}  + \lambda_{kps}\mathcal{L}_{kps}+ \lambda_{fit}\mathcal{L}_{fit}.
\vspace{-3mm}
\end{equation}

\begin{figure*}[t]
\begin{center}
\includegraphics[width=12cm, height=4cm]{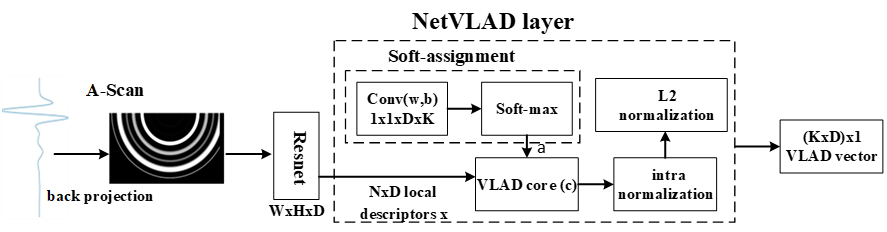}
\end{center}
\vspace{-6mm}
\caption{The GPRNet multi-task framework takes sparse point cloud data as input and simultaneously outputs segmentation results and the completed point clouds.}
\vspace{-5mm}
\label{descriptor}
\end{figure*}

\vspace{-3mm}
 \subsection{Underground Utilities Reconstruction} 
 \label{point}
 \vspace{-1mm}

Section \ref{para} describes how to extract sparse cross-sectional point clouds of objects from keypoints in B-scans. However, to generate a complete sparse point cloud of underground objects, we need to combine the sparse cross-sectional point clouds provided by ParabolaNet with pose information. For simulated data, the pose information can be directly obtained during data collection, while for real data, we calculate the pose using the method introduced in Section \ref{data}.
Due to the extreme sparsity of the point cloud, it is difficult to clearly distinguish the contours of objects. Therefore, we need to further complete the sparse point cloud map to recover the full structure of the objects. However, because of the complex underground conditions and the mixed distribution of different types of objects, directly applying point cloud completion methods may cause adjacent objects to blend together, leading to reconstruction errors.
Inspired by the works on 3D point cloud completion and segmentation \cite{huang2020pf, qi2017pointnet}, we propose GPRNet to address these challenges. We first perform point cloud segmentation on the sparse point cloud of underground objects and then complete the segmented objects to achieve a comprehensive 3D reconstruction of the underground environment.
Next, we propose a task learning network that effectively handles both data segmentation and completion. The core idea is to use a single encoder-multiple decoder architecture, where the encoder is responsible for extracting rich global features from the raw point cloud data, and two specialized decoders handle the segmentation and completion tasks, respectively. Our network takes sparse point cloud data as input and simultaneously outputs segmentation results and the completed point cloud.
Since we need to segment the point cloud and complete objects of different categories, relying solely on global features is insufficient; we also need to include the features of each point (local features). Therefore, in the encoder, we combine both local and global features. For the segmentation task, we use an MLP (multi-layer perceptron) to extract features based on this combination, which can be seen as a fusion of local and global features. For the completion task, we employ a multi-level structure within the network to process and generate point cloud features at different scales. High-level features typically capture the global structure, while low-level features focus on the local structure. Through layer-by-layer linear transformations and feature connections, features from different levels can propagate and integrate with each other.

 Encoder: Our encoder is inspired by the PointNet architecture, and we aim for the encoder to learn a representation capability that ensures the feature representation of point cloud data remains invariant after rotations and translations. We employ a T-Net module to predict the affine transformation matrix, performing affine transformations on the input to meet the requirements of transformation invariance. Using MLP, we extract N$\times$64 features from the input point cloud and insert another T-Net to predict the feature transformation matrix, performing a feature transform. N is the number of features. The aligned features are then further processed with MLP to obtain N$\times$1024 features, which are aggregated via MaxPooling to derive a 1$\times$1024 global feature vector. Considering the subsequent tasks of segmentation and completion, the global feature does not possess sufficient semantic information to support the tasks of the decoder. Therefore, we expand the 1$\times$1024 dimensional global feature to match the dimensionality of the N$\times$64 local features, concatenating the local and global features to form the final output of the encoder as a $N \times 1088$ dimensional feature.
 
Point Cloud Segmention Decoder:
The encoder output is a combination of $n \times 64$ local features and a 1024-dimensional global feature, which are fused together as the input to the segmentation decoder. The decoder utilizes an MLP $(512, 512, 128)$ to learn from the $N \times 1088$-dimensional feature space, generating $n \times 128$ vectors. Then, a perceptron of size $(128, m)$ is used to classify the final features, thus completing the segmentation task.

Point Cloud Completion Decoder:
The decoder takes the global feature vector $V$ as input and aims to output a dense point cloud result represented by \(M \times 3\). The fully-connected decoder \cite{achlioptas2018learning} excels at predicting the global geometry of the point cloud. However, its network structure makes it challenging to extract complete local geometric features. To overcome this limitation, we design our decoder as a hierarchical structure based on feature points \cite{huang2020pf}. In the main decoder layer, the global feature vector \(V\) passes through fully-connected layers to obtain three feature layers \(FC_i\) (\(FC_i := 1024, 512, 256\), for \(i = 1, 2, 3\)). Each feature layer is responsible for predicting point clouds at different resolutions. By concatenating features from three layers at different scales and employing fully-connected and linear transformations, a dense point cloud output is generated from our multi-resolution decoder. Benefitting from this multi-scale generating architecture, high-level features influence the expression of low-level features, and low-resolution feature points can propagate local geometric information to high-resolution predictions. Our experiments demonstrate that the point cloud completion decoder's predictions have fewer distortions and can retain the detailed geometry of the original missing point cloud.

\subsection{Mapping and Localization}
\vspace{-1mm}
In the field of subterranean structure detection, B-scan data furnish insights into the reflectivity profiles of subterranean objects across cross-sections, thereby delineating the contours of object surfaces. Each B-scan datum reflects the linear distance from the radar to the top surface of an object, that is, the depth information at the apex of the object. Conventionally, Ground-Penetrating Radar mapping has relied on predefined paths or GPS recordings to pinpoint the location of subterranean data collection. However, complex environments or those with limited GPS accessibility often pose constraints on the use of GPR. By equipping GPR vehicles with cameras and employing the VINS-Mono monocular SLAM framework, we calculate pose data during the B-scan scanning process, offering an efficient, accurate, and flexible solution for GPR mapping. Despite the absence of GPS positioning, SLAM technology enables the inference of device movement through continuous observational data and prior pose estimations, thus providing precise locational and directional information for B-scan data.
In Section \ref{para}, leveraging Parnet allows us to detect parabolic vertices within B-scan imagery, which represent the depths of subterranean objects. By integrating pose data computed via SLAM, we can concatenate these cross-sectional depth data, forming a sparse point cloud that depicts the topographical contours of subterranean objects.
In the context of using SLAM to estimate the pose of a GPR detector, the process of converting depth information, $d_i$, into three-dimensional spatial coordinates requires the integration of the detector's pose estimation. 
\vspace{-2mm}
\begin{equation}
(x, y, z) = R_t \cdot f(d_i, \text{angle}, \text{orientation}) + P_t
\label{pose1}
\vspace{-1mm}
\end{equation}

$d_i$ represents the depth, \textit{angle} represents the scanning angle, \textit{orientation} represents the orientation of the detector. $R_t$ is the rotation matrix at time $t$, $P_t$ is the position vector at time $t$. $(x_i, y_i, z_i)$ is the three-dimensional coordinate point in the global coordinate system.

Equation \ref{pose1} indicates that the pose information estimated by SLAM can transform the depth of each point extracted from GPR data into specific positions in the global reference frame. In this scenario, the Ground Penetrating Radar is configured to be horizontally placed and to scan vertically downward beneath the ground, with the detector's orientation pointing directly forward. Due to this configuration of the GPR, we can simplify the process of converting data into three-dimensional spatial coordinates. Each depth point \(d_i\) extracted from the B-scan data can be directly converted into coordinates within three-dimensional space. Since the radar scans vertically downward, the scanning angle can be considered to be around 0 degrees, meaning that the depth \(d_i\) directly corresponds to the point's position in the vertical (\(z\)) direction. Therefore, for each point, its local three-dimensional coordinates can be represented as \((0,0,d_i)\), where the \(x\) and \(y\) coordinates are 0 because the radar scans directly downward with no lateral shift. Thus, Eq. \ref{pose1} can be simplified as follows:
\vspace{-1mm}
\begin{equation}
    (x_i, y_i, z_i) = R_t \cdot (0,0,d_i) + P_t
    \vspace{-1mm}
\end{equation}

Each depth $d_i$ extracted from the B-scan data can be directly converted into coordinates in three-dimensional space. Since the radar scans vertically downward, the scanning angle can be considered to be 0 degrees, which means that the depth $d_i$ directly corresponds to the point's position in the vertical ($z$) direction. Therefore, for each point, its local three-dimensional coordinates can be represented as $(0,0,d_i)$, where the $x$ and $y$ coordinates are 0 because the radar scans directly downward with no lateral offset.
Through the above process, we can extract points representing the top contours of underground objects from each B-scan data and transform these points into the global coordinate system, constructing a global sparse point cloud:
\vspace{-2mm}
\begin{equation}
C_{\text{sparse}} = \bigcup_{i=1}^{N} (x_i, y_i, z_i)
\vspace{-2mm}
\end{equation}
\(C_{\text{sparse}}\) represents the global sparse point cloud.
After obtaining the sparse point cloud \(C_{\text{sparse}}\), we employ the Section \ref{point} point cloud completion network described in the section to fill in the sparse point cloud, resulting in a dense point cloud that yields a more complete representation of the underground structure.
Although A-scan data fundamentally differs from RGB data, the essence of A-scan captures the amplitude of electromagnetic energy, inherently embodying characteristics of subsurface information. Through the Back Projection (BP) algorithm, it is possible to transform energy of varying amplitudes into semi-spherical shapes at different depths. As illustrated in Fig. \ref{descriptor}, back projection is initially used to transform raw A-scan data into images with more distinct features, facilitating subsequent feature extraction.
\vspace{-2mm}
\begin{equation}
V(j,k) = \sum_{i=1}^N
    \frac{e^{\mathbf{w}_k^T \mathbf{x}_i + b_k}}{\sum_{k'}{e^{\mathbf{w}_{k'}^T \mathbf{x}_i + b_{k'}}}}
    \left( x_i(j) - c_k(j) \right),
\label{eq:vladlayer}
\vspace{-1.5mm}
\end{equation}
\noindent where $\{\mathbf{w}_k\}$, $\{b_k\}$ and $\{\mathbf{c}_k\}$ are sets of trainable parameters for each cluster $k$.
The NetVLAD layer aggregates the first order statistics of residuals $(\mathbf{x}_i - \mathbf{c}_k)$ in different parts of the descriptor space weighted by the soft-assignment $\bar{a}_k(\mathbf{x}_i)$ of descriptor $\mathbf{x}_i$ to cluster $k$. As illustrated in Fig. \ref{descriptor}, the input to this model is A-scan data processed through back projection. The NetVLAD layer signifies the commencement of the soft-assignment process. Therefore, the soft-assignment of the input array of descriptors $\mathbf{x}_i$ into $K$ clusters can be viewed as a two-step process: Initially, features are extracted from the input using a ResNet network, followed by a convolution operation with a set of $K$ filters $\{\mathbf{w}_k\}$ of $1 \times 1$, producing the output $s_k(\mathbf{x}_i) = \mathbf{w}_k^T \mathbf{x}_i + b_k$; Subsequently, the convolution output is processed through a softmax function $\sigma_k$, resulting in the final soft-assignment $\bar{a}_k(\mathbf{x}_i)$. This process weights the different terms in the aggregation layer. After normalization, the output is a $(K\times D) \times 1$ descriptor, aiding in feature matching for subsequent localization tasks.

\begin{figure*}[t]
\begin{center}
\includegraphics[width=12cm, height=4cm]{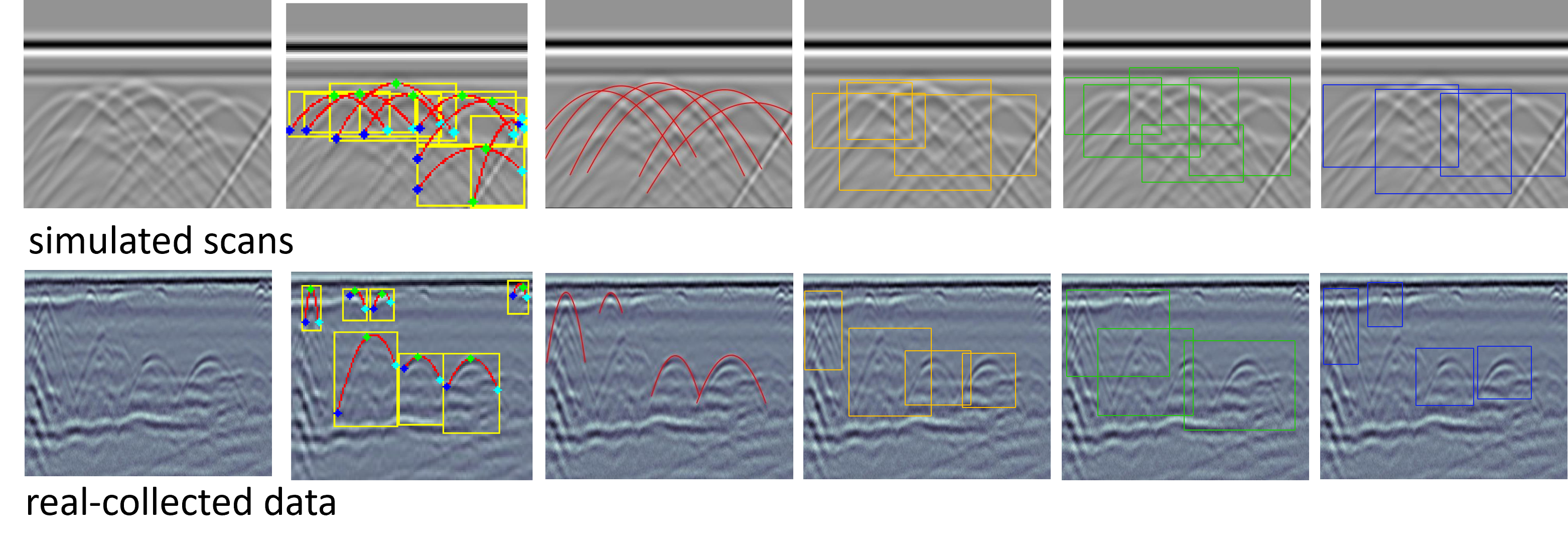}
\end{center}
\vspace{-8mm}
\caption{The qualitative results of the parabola detection compared with other detection methods are divided into two categories: simulated scans and real-collected data. From left to right, the sequence is: Raw created B-scans; our Parnet, \cite{lu20213d}, \cite{liu2020detection}, \cite{park2021improvement}, and \cite{feng2020gpr}. }
\vspace{-6mm}
\label{parnet_result}
\end{figure*}

\section{EXPERIMENT}
\vspace{-1mm}
In this section, we will subsequently describe the experimental evaluation metrics, implementation configuration, and both qualitative and quantitative results and comparisons, including our ParNet, GprNet, mapping, and localization. Analysis of ablation studies is also provided at the end.
\vspace{-2mm}
\subsection{Datasets}
We randomly combined spherical, rectangular, and cylindrical objects, varying their size, quantity, and positions, to generate a large number of underground 3D structures with different distributions. Using the gprMax software, we generated corresponding A-scan and B-scan images for training the ParNet network and for localization tasks. The real point cloud data were constructed based on the underground 3D structures. The sparse point cloud, formed by concatenating the keypoints detected in the B-scan images, was used as the input for the point cloud completion model.

\vspace{-1mm}
\subsection{GPR Parabola Signal Detection}
\vspace{-1mm}
As illustrated in Fig. \ref{parnet_result}, we evaluated the detection performance across both simulated and real Ground Penetrating Radar scans. It is noteworthy that our ParNet can accurately detect and precisely fit parabolic shapes. The bounding boxes produced by \cite{feng2020gpr} are fewer in number compared with the truth. Although \cite{liu2020detection} and \cite{park2021improvement} show improvement in terms of missing root targets, their bounding boxes are loosely defined with errors and overlaps in estimations. In contrast, our method achieves tight and accurate parabolic curves, moving beyond the rough rectangular bounding boxes. From Fig. \ref{parnet_result}, we observe that while \cite{lu20213d} achieves relatively correct parabolic fits, it misses faint and smaller parabolas and does not precisely align with the parabolas in the B-scans.

\begin{figure*}[t]
\begin{center}
\includegraphics[width=12cm, height=4cm]{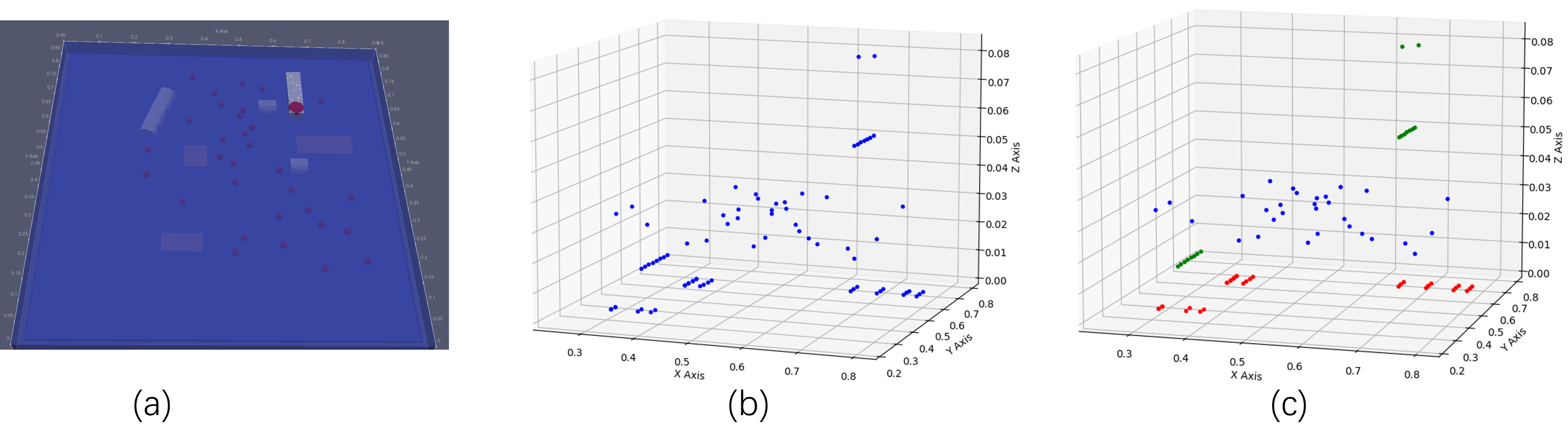}
\end{center}
\vspace{-6mm}
\caption{Point cloud segmentation results: (a) is ground truth; (b) shows the sparse point cloud obtained through ParNet; (c) depicts the point cloud segmentation results, segmenting underground objects such as stones, wooden boards, and cylindrical items. }
\vspace{-4.5mm}
\label{segment_result}
\end{figure*}

\begin{figure*}[t]
\begin{center}
\includegraphics[width=12cm, height=6cm]{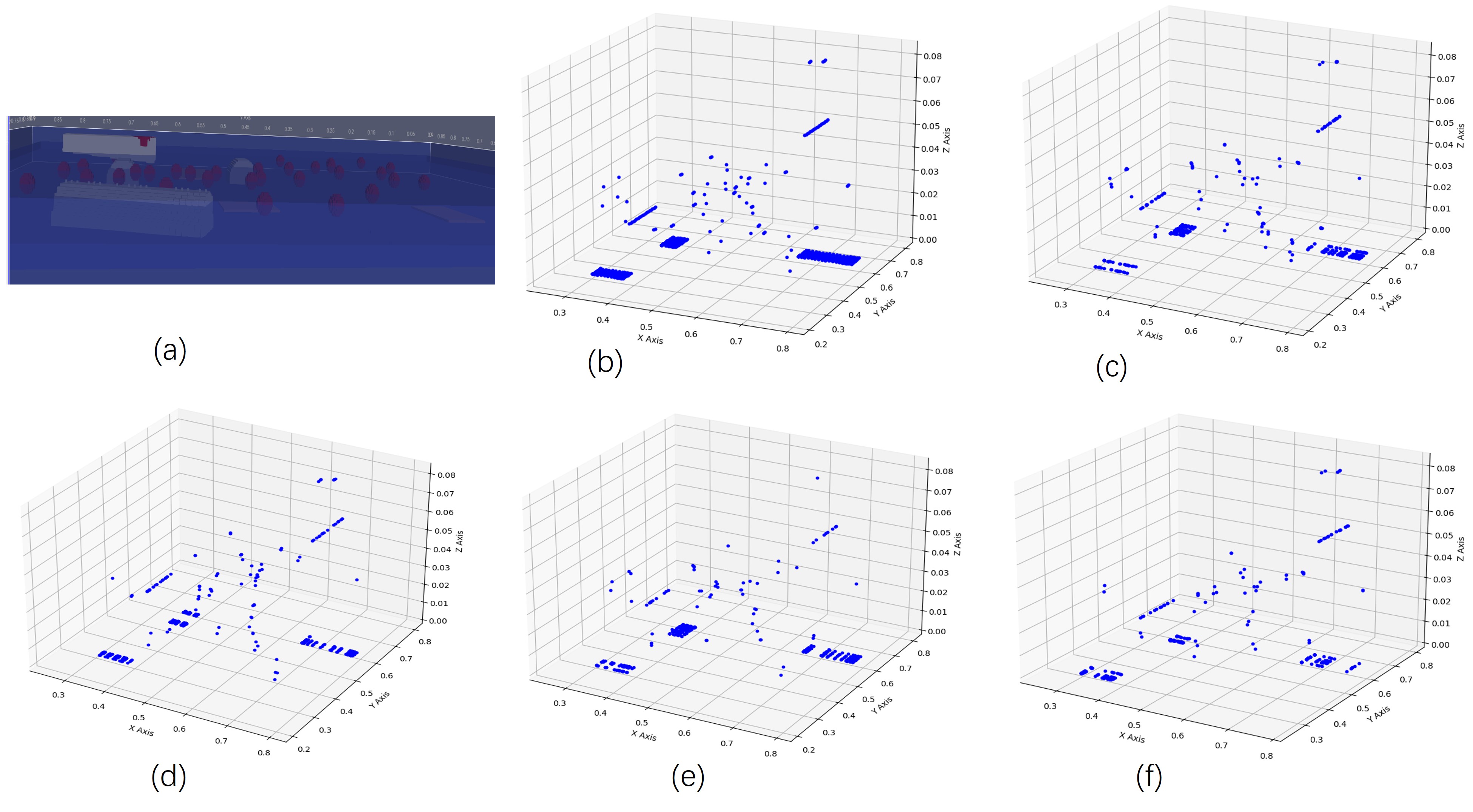}
\end{center}
\vspace{-6mm}
\caption{The comparison of completion results between other methods and our network. From left to right: (a) the ground truth; (b) our result,(c) result from PCN \cite{yuan2018pcn}; (d) result from GRNet \cite{xie2020grnet};
(e) result from PF-Net \cite{huang2020pf}; (f) result from Wang et al \cite{wang2020point}. }
\vspace{-5mm}
\label{complete}
\end{figure*}

\vspace{-1mm}
\subsection{Underground Scene Mapping}
\vspace{-1mm}
Utilizing the sparse point cloud obtained from ParNet, we employ GPRNet for point cloud completion and segmentation, achieving underground mapping. Fig. \ref{segment_result} shows the segmentation results, with (b) representing the input sparse point cloud. Our network segments stones, wooden boards, and cylinders using different colors, producing segmentation results that closely match real-world conditions. 
Fig. \ref{complete} illustrates the differences between our point cloud completion results and those from other methods. Completing complex underground sparse point clouds directly poses a challenge, as small stones cause significant interference. Additionally, due to the sparseness of the point cloud, if the network fails to recognize which points belong to the same object, it is likely to complete them as two separate objects, thus introducing errors in underground mapping. To address this, we introduce a segmentation task that allows the network to complete point clouds of single objects without causing fragmentation. Compared to the real situation in (a), our results closely match the ground truth.
To quantitatively analyze the accuracy of our mapping, we use the evaluation metrics from \cite{gadelha2018multiresolution,lin2018learning}, which include Pred-GT (prediction to ground truth) and GT-Pred (ground truth to prediction) errors. Pred-GT error calculates the average squared distance from each predicted point to its nearest ground truth point, assessing the prediction's deviation from the ground truth. Conversely, GT-Pred error measures the average squared distance from each ground truth point to its nearest predicted point, reflecting the extent to which the predicted shape covers the ground truth surface.
Table \ref{comp_table} demonstrates that our method outperforms others in all categories for both Pred-GT and GT-Pred errors. It is important to note that the overall error in the completed map originates from two factors: the prediction error in the missing area and changes to the original partial shape. Since our method takes the partial shape as input and only completes the missing areas, it does not alter the original partial shape. To ensure a fair evaluation, we also calculate Pred-GT and GT-Pred errors specifically for the missing areas. The Chamfer distance and F-Score results in Table \ref{comp_table} indicate that our mapping accuracy surpasses that of other methods.

\begin{table}[t]
\centering
\begin{tabular}{l|l|l|l|l}
\hline
                                                 & CD   & \begin{tabular}[c]{@{}l@{}}Pred-GT\end{tabular} & \begin{tabular}[c]{@{}l@{}}GT-Pred\end{tabular} & F-Score \\ \hline
PCN \cite{yuan2018pcn}          & 3.14 & 1.33                                               & 1.27                                               & 0.61    \\
GRNet \cite{xie2020grnet}       & 4.57 & 1.81                                               & 1.59                                               & 0.49    \\
PF-Net \cite{huang2020pf}       & 3.49 & 1.45                                               & 1.38                                               & 0.57    \\
Wang \cite{wang2020point} & 4.21 & 1.69                                               & 1.62                                               & 0.42    \\
Ours                                             & 1.45 & 0.72                                               & 0.68                                               & 0.87    \\ \hline
\end{tabular}
\caption{Point completion results of our method compared with other methods using Chamfer Distance (CD) with L2 norm. CD, Pred-GT, and GT-Pred are lower the better. F-Score is the higher the better.}
\label{comp_table}
\vspace{-6mm}
\end{table}

\begin{table}[t]
\vspace{2mm}
\centering
\begin{tabular}{l|lll}
\hline
material         & \begin{tabular}[c]{@{}l@{}}Relative\\ Permittivity\end{tabular} & Conductivity & Recall \\ \hline
sand      & 3                                                               & 0.02         & 88\%   \\

clay        & 10                                                              & 0.3          & 84.5\% \\

loamy     & 20                                                              & 0.5          & 81     \\ \hline
\end{tabular}
\caption{Localization recall results for six different types of soil. }
\label{material}
\vspace{-10mm}
\end{table}

\vspace{-2mm}
\subsection{Localization}
\vspace{-1mm}
Through feature matching of A-scan data, we identify the specific locations of unknown sites on an established map. Initially, we collected A-scan data from 2000 different locations within a broad area, representing known positions on the map. Each A-scan dataset captures the underground features of its corresponding location, laying the groundwork for subsequent matching processes. We randomly select or designate an unknown site and collect A-scan data from it. Then, using a feature matching algorithm, we compare the A-scan data of this unknown site with that of the 2000 known locations in the map database. Using the NetVLAD network, we extract underground image features from the A-scan data and apply matching algorithms to find the database location most similar to the A-scan features of the unknown site. We assume the unknown site is closest to the location corresponding to the matched A-scan, serving as our localization result. 
To evaluate the effectiveness of this localization method, we introduce recall rate as the primary evaluation metric. The recall rate is calculated by comparing the positions matched by the algorithm with the actual real positions of the unknown sites. Specifically, if the location identified by the matching algorithm is the closest among all known locations to the unknown site, we consider the match successful. A total of 200 A-scan datasets from unknown locations were used for position identification. 
Moreover, acknowledging the influence of varying soil compositions on A-scan data representation, we extended our experiment to include six distinct soil types: dry sand, wet sand, dry clay, wet clay, loamy sand, and wet loamy sand, each associated with unique Relative Permittivity and Conductivity values. The variances in these properties notably impact the radar wave's velocity and the signal's attenuation rate, thereby affecting the A-scan results' reflectivity and depth detectability. 
Table \ref{material} shows a notably high matching accuracy for lower conductivity and relative permittivity materials, such as sand, reaching 88\%. Although accuracy slightly reduced for clay to 84.5\%, and further to 81\% for loamy sand variants, these results still fell within a satisfactory accuracy range. Through this extensive evaluation across varied geological conditions, we aim to delve into the nuanced effects of soil composition on A-scan feature matching accuracy, thereby enriching our understanding and methodologies for subterranean detection and localization.

\begin{table}[t]
\vspace{-3mm}
\begin{tabular}{l|l|l|l|l|l|l}
\hline
T-net & \begin{tabular}[c]{@{}l@{}}segmentation decoder\end{tabular} & \begin{tabular}[c]{@{}l@{}}global+local\end{tabular} & CD   & \begin{tabular}[c]{@{}l@{}}Pred-GT\end{tabular} & \begin{tabular}[c]{@{}l@{}}GT-Pred\end{tabular} & F-Score \\ \hline
      &                            &                                & 5.79 & 2.12                                               & 1.74                                               & 0.35    \\
   \checkmark   &                         &                          & 4.23 & 1.92                                               & 1.49                                               & 0.47    \\
 \checkmark    &            \checkmark                        &              & 2.29 & 1.05                                               & 0.89                                               & 0.72    \\
   \checkmark   &              \checkmark                &       \checkmark                 & 1.45 & 0.72                                               & 0.68                                               & 0.87    \\ \hline
\end{tabular}
\caption{The effect of the T-net, segmentation decoder, Aggregating Local and Global feature. }
\vspace{-9mm}
\label{Ablation_result}
\end{table}

\vspace{-2mm}
\subsection{Ablation Studies}
\vspace{-1mm}
In Table \ref{Ablation_result}, we analyze the impact of different modules on mapping accuracy. The results indicate that the T-net, segmentation decoder, and aggregation of local and global features significantly contribute to generating a complete and dense underground point cloud map. With the combined contributions of these three optimizations, compared to the baseline without them, the Chamfer Distance (CD) decreased from 5.79 to 1.45, pred-GT from 2.12 to 0.72, GT-pred from 1.74 to 0.68, and the F-Score increased to 0.35.
\vspace{-2mm}
\section{CONCLUSION}
This paper introduces a non-destructive underground mapping and localization framework based on GPR. The network ParNet, specifically designed for B-scan parabola detection, accurately identifies keypoints from GPR scans and uses them for curve fitting to determine the parabola's apex. The reconstruction of the top contours of underground objects does not require prior knowledge of object shapes for detection. GPRNet employs segmentation and completion decoders to effectively perform point cloud completion, even in complex scenarios. The completed underground maps can be used for localizing unknown locations using A-scan data. Extensive experiments demonstrate that the proposed method achieves superior GPR signal detection and underground object reconstruction. It surpasses existing state-of-the-art methods. Localization based on the reconstructed underground maps has an error margin within 20\%, meeting current demands for accuracy in localization.

\clearpage  

%
%
\bibliographystyle{splncs04}
\bibliography{main}
\end{document}